%% file: main.tex
\pdfoutput=1

\documentclass[11pt]{article}
\usepackage{authblk}

\usepackage{EMNLP2023}

\usepackage{times}
\usepackage{latexsym}

\usepackage[T1]{fontenc}

\usepackage[utf8]{inputenc}

\usepackage{microtype}

\usepackage{inconsolata}

\usepackage{array}
\usepackage{multirow}
\usepackage{graphicx}
\usepackage[justification=centering]{caption}
\usepackage{amsmath,amssymb}
\usepackage[ruled,vlined]{algorithm2e}
\usepackage[capitalize]{cleveref}
\usepackage{color, colortbl}
\usepackage[dvipsnames]{xcolor}

\definecolor{Gray}{gray}{0.9}
\definecolor{Red}{RGB}{255, 158, 158}
\definecolor{DirectScoring}{RGB}{255, 242, 207}
\definecolor{PairwiseScoring}{RGB}{217, 220, 250}


\newcolumntype{H}{>{\setbox0=\hbox\bgroup}c<{\egroup}@{}}

\title{Direct-Scoring NLG Evaluators Can Use Pairwise Comparisons Too}

\author[1]{Logan Lawrence}
\author[2]{Ashton Williamson}
\author[2]{Alexander Shelton}
\affil[ ]{$^1$University of Massachusetts Amherst, $^2$School of Computing, Clemson University}
\affil[ ]{\texttt{lclawrence@umass.edu, \{taw2,alshelt\}@clemson.edu}}

\begin{document}
\maketitle

\begin{abstract}
    As large-language models have been increasingly used as automatic raters for evaluating free-form content, including document summarization, dialog, and story generation, work has been dedicated to evaluating such models by measuring their correlations with human judgment. For \textit{sample-level} performance, methods which operate by using pairwise comparisons between machine-generated text perform well but often lack the ability to assign absolute scores to individual summaries, an ability crucial for use cases that require thresholding. In this work, we propose a direct-scoring method which uses synthetic summaries to act as pairwise machine rankings at test time. We show that our method performs comparably to state-of-the-art pairwise evaluators in terms of axis-averaged sample-level correlations on the SummEval (\textbf{+0.03}), TopicalChat (\textbf{-0.03}), and HANNA (\textbf{+0.05}) meta-evaluation benchmarks, and release the synthetic in-context summaries as data to facilitate future work.
\end{abstract}

\section{Introduction}

As large-language models (LLMs) continue to push the state-of-the-art in natural-language generation (NLG), the task of evaluating the quality of their outputs has become an increasingly complex challenge. Traditional approaches to natural language evaluation that compare n-gram overlap between source and reference text samples \cite{papineni2002bleu, rouge, meteor}, despite their broad popularity, largely fail to capture semantic information. Model-based evaluation metrics built on smaller pretrained language models \cite{bertscore, bartscore, moverscore, feqa, qags, mqag, qafacteval, qafacteval, apes, chen_semantic, questeval} have offered a more robust solution to NLG evaluation, but often struggle in challenging evaluation scenarios \cite{he-etal-2023-blind, hanna2021fine} and have failed to achieve strong alignment with human judgment on modern meta-evaluation benchmarks \cite{summeval_dataset}. 

To overcome these limitations, recent work has explored the viability of using LLMs for NLG evaluation in a prompting-based scenario, enabling their use as both zero-shot and reference-free evaluators \cite{liu-etal-2023-g, zepo, pairs2024}. Of these, comparison-based approaches \cite{gao2025evaluating} have seen considerable attention in recent work, owing largely to their demonstrated superior alignment with human judgment \cite{liusie2024llm}. 


However, the relative judgment of comparison-based approaches limits their applicability to a number of common use cases, specifically threshold-based scenarios that require an absolute score for filtering and sorting. As an alternative, we propose a direct-scoring evaluator which \textit{uses pairwise comparisons with synthetic samples} and show that it performs comparably to comparison-based approaches over the SummEval \cite{summeval_dataset}, TopicalChat \cite{topicalchat_dataset}, and HANNA \cite{hanna_dataset} meta-evaluation datasets. Finally, we publish all work: the synthetic summaries for each dataset over a variety of LLMs, code and prompts needed to generate the summaries, as well as evaluation utilities.\footnote{\url{https://github.com/Racerx250/direct_scoring_synthetic_pairwise}}

\input{figures/preview_combined}
\section{Methodology}
Next, we describe the proposed method (depicted in \cref{fig:preview_combined}) which consists of two pieces: (1) the creation of synthetic in-context examples of various qualities, (2) the inference setup given these generated examples. 
\subsection{Creating Synthetic In-Context Examples}
Given an LLM, task, context, and quality dimension, we wish to generate $N$ responses of varying quality-levels for the context. For example, we can take the "task" to be summarization, "context" to be news articles, and "responses" to be summaries. Specifically, we want to generate summaries with \textit{inter-rating} consistency: making sure that examples of different scores actually improve with respect to the dimension as rating increases.

We propose to generate examples of monotonically-increasing quality by first starting with the \textit{extremes} of the ratings, then prompting the LLM to generate examples of \textit{intermediate quality} between those previously existing. Specifically, letting $N=5$ (which matches the rating schemes of SummEval and NewsRoom \cite{summeval_dataset, newsroom_dataset}) we start by prompting the LLM for the worst and best possible summaries, i.e. scores $1$ and $5$:
\definecolor{quotecolor}{rgb}{0.87, 0.95, 0.96}
\begin{quote}
    \makebox[\linewidth]{%
        \colorbox{quotecolor}{%
            \hspace*{0mm} 
            \begin{minipage}{\dimexpr\linewidth+10\fboxsep\relax} 
                \fontsize{9pt}{10pt}\selectfont 
                \textbf{(Score 1 - Lowest)} What is the \textit{worst} possible summary of the following article with respect to \texttt{[quality]}, \texttt{[description]}? \\
                
                Article: \texttt{[article]}
            \end{minipage}%
        }%
    }
\end{quote}
\begin{quote}
    \makebox[\linewidth]{%
        \colorbox{quotecolor}{%
            \hspace*{0mm} 
            \begin{minipage}{\dimexpr\linewidth+10\fboxsep\relax} 
                \fontsize{9pt}{10pt}\selectfont 
                \textbf{(Score 5 - Highest)} What is the \textit{best} possible summary of the following article with respect to \texttt{[quality]}, \texttt{[description]}? \\
                
                Article: \texttt{[article]}
            \end{minipage}%
        }%
    }
\end{quote}

where \texttt{[quality]} is the plain-text name of a quality dimension ("Consistency", "Coherence", "Relevance", or "Fluency" in the case of SummEval \cite{summeval_dataset}) and \texttt{[description]} is an explanation of \texttt{[quality]}. Let \texttt{summary$_1$} and \texttt{summary$_5$} refer to the results of prompting the LLM as above. For the intermediate summaries \texttt{summary$_{2, 3, 4}$}, we propose to generate them recursively:
\vspace{-.1cm}
\begin{quote}
    \makebox[\linewidth]{%
        \colorbox{quotecolor}{%
            \hspace*{0mm} 
            \begin{minipage}{\dimexpr\linewidth+10\fboxsep\relax} 
                \fontsize{9pt}{10pt}\selectfont 
                \textbf{(Score $i \in {2, 3, 4}$ - Intermediate)} For the two summaries, what is a new summary of \textit{intermediate} \texttt{[quality]}, \texttt{[description]}? \\
                
                Article: \texttt{[article]}

                Worse Summary: \texttt{[summary$_{i - 1}$]}

                Better Summary: \texttt{[summary$_{i + 1}$]}
            \end{minipage}%
        }%
    }
\end{quote}
We provide the final full prompts used for each setting and score in \cref{sec:appendix:prompt_for_summaries_and_prediction} and also note that ablation over the prompt and amount of generated examples $N$ is performed in \cref{sec:performance}. From this point, we refer to the summaries as \texttt{icl$_{1,2,3,4,5}$}.

\subsection{Pairwise Probability Calculation on Synthetic Examples}\label{sec:method:probability_calculation}
\input{tables/main_results}

Given the synthetic examples \texttt{icl$_{1,2,3,4,5}$} for a given article and dimension, as well as a prospective machine-generated summary \texttt{machine}, we calculate the probability of the model responding with \texttt{"Better", "Worse", "Similar"}, and take the weighted sum over the synthetic scores. Note that this differs from previous direct scoring approaches like G-Eval \cite{liu-etal-2023-g}, which take the probabilities over scores themselves; we prompt each score \textit{pairwise} with a generated reference, then take the summation.  
\begin{quote}
    \makebox[\linewidth]{%
        \colorbox{quotecolor}{%
            \hspace*{0mm} 
            \begin{minipage}{\dimexpr\linewidth+10\fboxsep\relax} 
                \fontsize{9pt}{10pt}\selectfont 
                \textbf{(Score $i$ Evaluation)} Here's a news article, a reference summary, and prospective summary. How does the prospective summary compare to the reference with respect to \texttt{[quality]}, \texttt{[description]}? Respond only with "Worse," "Better," or "Similar."    \\
                
                Article: \texttt{[article]}

                Reference Summary: \texttt{[icl$_{i}$]}

                Prospective Summary: \texttt{[machine]}
            \end{minipage}%
        }%
    }
\end{quote}

Given the above prompt for score $i$, we calculate and sum the log probabilities of the LLM responding with \texttt{"Worse"}, \texttt{"Better"}, or \texttt{"Similar"}. Next, we calculate the softmax over these summed log probabilities, which we refer to as $p(\texttt{"Worse"}|i)$, $p(\texttt{"Better"}|i)$, and $p(\texttt{"Similar"}|i)$. Using these probabilities, we construct final prediction by taking weighted average of scores:
\begin{align*}
    s(\cdot) &= \sum_{i \in 1,...,5} [i, -i, 0] * \begin{bmatrix}
        p(\texttt{"Better"}|i) \\
        p(\texttt{"Worse"}|i) \\
        p(\texttt{"Similar"}|i) \\
    \end{bmatrix}
\end{align*}

\section{Experiments}

\paragraph{Datasets} We evaluate our method on three meta-evaluation benchmarks: SummEval \cite{summeval_dataset} for summarization, TopicalChat \cite{topicalchat_dataset} for dialog, and HANNA \cite{hanna_dataset} for story generation. Contrary to prior works, we do not report NewsRoom \cite{newsroom_dataset} in the main results due to a surprising finding that BERTScore \cite{bertscore} with \texttt{roberta-large} attains high performance (see \cref{sec:appendix:bertscore_and_newsroom}).

\paragraph{Metric} Following previous works \cite{zepo, zhong2022towards, fu2023gptscore}, we evaluate the efficacy of our method using \textit{sample-level} correlations with machine generated summaries. Namely, for a given correlation metric (e.g. Spearmans $\rho$) and axis, a group of abstractive summarization machines of size $M$, and a number of full-texts $T$, we calculate the following metric:
\begin{align*}
    F^{sample} &= \frac{1}{n}\sum^{T}_{i=1}\rho(\begin{bmatrix}
        pred_{i,1}, human_{i,1} \\
        ... \\
        pred_{i,M}, human_{i,M}
    \end{bmatrix}) 
\end{align*}
where $pred_{i,j}$ is the prediction for the $j^{th}$ machine's response on the $i^{th}$ document and $human_{i,j}$ is the ground truth label for that machine response. We provide a brief discussion of the aggregation metrics, namely the originally proposed \textit{system-level} \cite{summeval_dataset} and \textit{summary-level} metrics in \cref{sec:appendix:all_performances}.

\paragraph{Baselines} We provide direct-scoring baselines G-Eval \cite{liu-etal-2023-g}, a direct-scoring baseline ("Scoring") which uses the same prompt as G-Eval, as well as classical metrics such as BERTScore \cite{bertscore} and GPTScore \cite{fu2023gptscore}. For pairwise baselines, we provide ZEPO \cite{zepo} and PairS-beam \cite{pairs2024}. For all sampling-based generation, we conduct all experiments with the same hyperparameters.

\section{Results}\label{sec:performance}

\paragraph{Main Results}\label{sec:performance:main_performance}

We show the sample-level performance of our method in \cref{table:main_results}. In terms of average sample-level correlation across the axes of each dataset, our method performs the best on SummEval ({\color{Green} \textbf{+0.03/+0.01}} for Mistral 7B \cite{jiang2023mistral7b} / Llama-3 8B \cite{llama3} versus next highest performer), second best on TopicalChat ({\color{BrickRed} \textbf{-0.08/-0.03}}), and best on HANNA ({\color{Green} \textbf{+0.01/+0.05}}). However, we do not find that our method is always better over axes, often being the first or runner-up. 

\paragraph{Performance of Different Prediction Methods} In the first section of \cref{table:prediction_ablation}, we ablate on the prediction method used to generate final scores for a given text. "Sample" refers to sampling $n$ responses using constrained decoding and "$p(\texttt{"Yes"}), p(\texttt{"No"})$" refers to using \texttt{"Yes"} and \texttt{"No"} with a comparison-based prompt (see \cref{sec:appendix:prompt_for_summaries_and_prediction}). Firstly, we find that using a comparative-based prompt in our setup increases performance over the proposed setting. Next, we find that increasing $n$ results in monotonically increasing performance, with the maximum achieved at $n=1000$ with \textbf{36.44} sample-level correlation. We also note that this is still below the performance of our method (\textbf{37.43} -  $p(\texttt{"Better"}), p(\texttt{"Worse"}),\dots$), notably a gap of \textbf{0.99}. This indicates that probability generation is essential to our method, which we fix for the second section of the table. 

When varying the amount of examples to compare to at prediction time (second section of \cref{table:prediction_ablation}), we see that increasing the amount of examples, even past $N = 5$, increases performance with the maximum achieved by $N = 9$ (\textbf{37.80} vs. \textbf{37.43}). However, we still keep $N = 5$ for simplicity and its alignment with popular meta-evaluation datasets.

\paragraph{Performance Across Architectures}\label{sec:performance:architecture_performance} In the first section of \cref{table:architecture_performance}, we show the performance of our method when using different LLMs for prediction after generating synthetic examples. We find that the performance of our method can be further boosted by using more powerful backbones, as in the vanilla Direct Scoring setting, models with greater all-around performance (as measured by MMLU $\uparrow$) do not correspond to better-performing evaluators, but under our method this discrepancy is fixed. 

\input{tables/method_ablation}
\input{tables/direct_architecture_system_sample}

Similarly, when using different LLMs for synthetic example generation (second section of \cref{table:architecture_performance}), we find performance increases with more powerful models, albeit less consistently. Most notably, OLMo-2-7B \cite{olmo20242} underperforms relative to its MMLU score, producing less useful summaries than Mistral-v0.1-7B.

\section{Conclusion}
In this work, we proposed an LLM-based direct-scoring evaluation framework which uses synthetic in-context examples from LLMs to assign absolute scores to machine-generated summaries. We found that the method produces comparable average sample-level correlations to comparison-based approaches. We ablated on the method to find that probability generation is essential to performance, increasing the granularity of examples moderately boosts performance, and that LLMs with more instruction-following ability are higher performing with our method. This method addresses the need for a direct scoring metric with performance comparable to that of state-of-the-art comparison-based approaches, allowing for use cases involving thresholding. We also publicly release the synthetic summaries for further work.

\section{Limitations}
\paragraph{Choice of LLMs} We ablate on on a selection of small-sized (between 7 and 8 billion parameters) LLMs to compare to prior work and to adhere to our computational budget. Behavior for this subset of models may not extrapolate to larger LLMs. We leave investigations into the scalability of our approach with regard to model size to future work. 

\paragraph{Computational Cost} Our proposed approach requires $N$ synthetic reference examples to be generated per source article and task (quality dimension), plus an additional $N$ inputs to the LLM to compare the input text to each of the reference examples. This could become computationally prohibitive for scenarios in which there is no repetition in source articles, especially for larger values of $N$, resulting in both reduced inference speed and higher financial cost. Given that our ablation studies (see Table \ref{table:prediction_ablation}) show reduced performance at lower values of $N$, future work into both increasing synthetic summary generation efficiency and improving performance at lower values of $N$ may be useful.

\bibliographystyle{bibliography/acl_natbib}
\bibliography{bibliography/bib}

\appendix

\input{appendix}

\end{document}

%% file: figures/preview_combined.tex
\begin{figure*}[ht!]
    \centering
    \captionsetup{type=figure}
    \includegraphics[width=\linewidth]{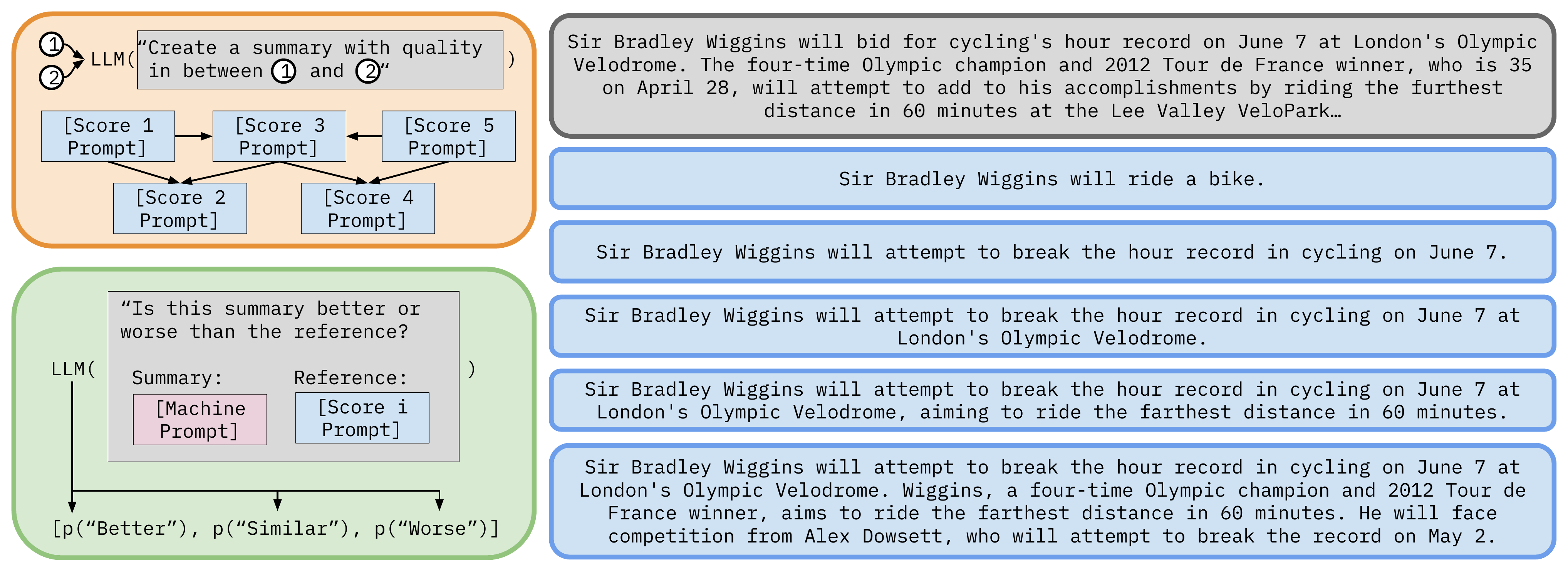}
    \vspace{-20pt}
    \captionof{figure}{\textbf{Overview of Method.} \textbf{(Left)} First an LLM is prompted to generate summaries reflecting various levels of quality using a contrastive scheme (orange), then compared to the machine summary to generate probabilities over comparative language (green), eg. \texttt{"Better", "Similar",} and \texttt{"Worse"}. \textbf{(Right)} We show an example of summaries (blue) of increasing quality (scores $1$ through $5$) for a SummEval article (grey) on the "consistency" column generated over the course of our method. }
    \label{fig:preview_combined}
\end{figure*}%

%% file: tables/main_results.tex
\begin{table*}[ht!]
\setlength{\belowcaptionskip}{-10pt}
\setlength{\tabcolsep}{3pt}
\begin{center}
{\small
\begin{tabular}{lccccccccccccccc}
\hline
 \\[-1.6ex]
 & \multicolumn{5}{c}{\textbf{SummEval}} & & \multicolumn{4}{c}{\textbf{TopicalChat}} & & \multicolumn{4}{c}{\textbf{HANNA}}\\[1ex]
\textbf{Method} \hspace{1.3cm} & \textbf{COH} & \textbf{CON} & \textbf{FLU} & \textbf{REL} & \textbf{AVG} & \hspace{0.2cm} & \textbf{NAT} & \textbf{ENG} & \textbf{OVE} & \textbf{AVG} & \hspace{0.2cm} & \textbf{COH} & \textbf{SUR} & \textbf{COM} & \textbf{AVG}\\
 \\[-2ex]
\hline
 \\[-2ex]
\rowcolor{Gray}
Other Metrics &&&&&&&&&&&&&&&\\
BERTScore$^1$ (F1) & 0.28 & 0.19 & 0.11 & \textbf{0.31} & 0.23 && \textbf{0.09} & \textbf{0.16} & \textbf{0.19} & \textbf{0.15 } && \textbf{0.25} & 0.22 & \textbf{0.30} & \textbf{0.26} \\
BERTScore$^2$ (F1) & \textbf{0.38} & \textbf{0.35} & \textbf{0.40} & 0.28 & \textbf{0.35} && 0.04 & -0.04 & 0.07 & 0.02 && \textbf{0.25} & 0.12 & 0.17 & 0.18 \\
GPTScore & 0.28 & 0.31 & 0.38 & 0.22 & 0.30 & & - & - & - & - && 0.22 & \textbf{0.25} & 0.08 & 0.18 \\
 \\[-2ex]
\hline
 \\[-2ex]
\rowcolor{Gray}
Mistral 7B &&&&&&&&&&&&&&& \\
\rowcolor{PairwiseScoring}
ZEPO & \underline{0.29} & 0.32 & 0.13 & \underline{0.30} & 0.26 && 0.14 & 0.25 & 0.28 & 0.22 && - & - & - & - \\
\rowcolor{PairwiseScoring}
PairS-beam & 0.28 & 0.31 & 0.18 & 0.24 & 0.25 && \textbf{0.41} & \textbf{0.41} & \underline{0.33} & \textbf{0.38} && 0.29 & \textbf{0.27} & 0.31 & 0.29 \\
\rowcolor{DirectScoring}
Direct Scoring & 0.23 & \underline{0.37} & \underline{0.19} & 0.19 & 0.25 && 0.26 & 0.17 & 0.32 & 0.25 && 0.30 & \underline{0.26} & 0.37 & 0.31 \\
\rowcolor{DirectScoring}
G-Eval & 0.25 & \textbf{0.39} & \textbf{0.20} & 0.25 & \underline{0.27} && 0.26 & 0.28 & \textbf{0.35} & \underline{0.30} && \textbf{0.34} & 0.25 & \underline{0.39} & \underline{0.33} \\
\rowcolor{DirectScoring}
Ours & \textbf{0.35} & 0.34 & 0.18 & \textbf{0.33} & \textbf{0.30} && \underline{0.31} & \underline{0.36} & 0.22 & \underline{0.30} && \underline{0.33} & 0.25 & \textbf{0.44} & \textbf{0.34} \\
 \\[-2ex]
\hline
 \\[-2ex]
\rowcolor{Gray}
Llama-3 8B &&&&&&&&&&&&&&& \\
\rowcolor{PairwiseScoring}
ZEPO & \underline{0.40} & 0.25 & \underline{0.30} & 0.39 & 0.34 && 0.16 & 0.26 & \underline{0.46} & 0.29 && - & - & - & - \\
\rowcolor{PairwiseScoring}
PairS-beam & 0.35 & \textbf{0.42} & \textbf{0.32} & 0.35 & 0.36 && \textbf{0.47} & \textbf{0.56} & 0.43 & \textbf{0.49} && 0.36 & 0.22 & 0.31 & 0.30 \\
\rowcolor{DirectScoring}
Direct Scoring & 0.35 & 0.32 & 0.23 & \textbf{0.46} & 0.34 && 0.33 & 0.32 & 0.40 & 0.35 && 0.26 & 0.17 & 0.32 & 0.25\\
\rowcolor{DirectScoring}
G-Eval & 0.34 & 0.29 & 0.22 & \underline{0.42} & 0.32 && 0.38 & 0.43 & \textbf{0.53} & 0.45 && \textbf{0.44} & \underline{0.29} & \underline{0.42} & \underline{0.38} \\
\rowcolor{DirectScoring}
Ours & \textbf{0.44} & \underline{0.41} & 0.26 & 0.36 & \textbf{0.37} && \underline{0.43} & \underline{0.50} & 0.43 & \underline{0.46} && \underline{0.42} & \textbf{0.37} & \textbf{0.49} & \textbf{0.43} \\
 \\[-2ex]
\hline
\end{tabular}
\caption{\textbf{Main Results.} Sample-level Spearman correlations of Llama-3 8B over the various axes of SummEval \cite{summeval_dataset}, TopicalChat \cite{topicalchat_dataset}, and HANNA \cite{hanna_dataset} for \colorbox{PairwiseScoring}{comparison-based evaluators} and \colorbox{DirectScoring}{direct-scoring evaluators}. "Average" refers to the mean of the columns for each dataset. Best performance is \textbf{bolded} and second-best is \underline{underlined}. BERTScore$^1$ refers to using \texttt{bert-base-uncased} whereas $^2$ refers to using \texttt{roberta-large}.}
\label{table:main_results}
}
\end{center}
\end{table*}

%% file: tables/method_ablation.tex
\begin{table}[ht!]
\setlength{\belowcaptionskip}{-10pt}
\setlength{\tabcolsep}{9pt}
\begin{center}
{\small
\begin{tabular}{lc}
\hline
 \\[-1.6ex]
\textbf{Method} & \textbf{Avg. SummEval}\\
 \\[-2ex]
\hline
 \\[-2ex]
\rowcolor{Gray}
Sampling & \\
Sample, $n = 1$ & 9.38 $\pm$ 0.67 \\
Sample, $n = 3$ & 14.68 $\pm$ 1.51 \\
Sample, $n = 10$ & 21.24 $\pm$ 2.26 \\
Sample, $n = 100$ & 30.91 $\pm$ 1.32 \\
Sample, $n = 1000$ & 36.44 $\pm$ 0.42 \\
$p(\texttt{"Better"}), p(\texttt{"Worse"}),\dots$ & 37.43 \\
$p(\texttt{"Yes"}), p(\texttt{"No"}),\dots$ & \textbf{38.60} \\
 \\[-2ex]
\hline
 \\[-2ex]
 \rowcolor{Gray}
\# of Examples &\\
$N = 2$ & 34.83\\
$N = 3$ & 36.58 \\
$N = 5$ & 37.43 \\
$N = 9$ & \textbf{37.80} \\
 \\[-2ex]
\hline
\end{tabular}
\caption{\textbf{Ablation on Prediction Method Used.} "Sampling" refers to having the LLM generate a pairwise judgment via constrained decoding, whereas $N$ refers to the amount of in-context examples used to compare to the target summary. "$\pm$" refers to the standard deviation over 5 trials.}
\label{table:prediction_ablation}
}
\end{center}
\end{table}

%% file: tables/direct_architecture_system_sample.tex
\begin{table}[ht!]
\setlength{\belowcaptionskip}{-10pt}
\setlength{\tabcolsep}{9pt}
\begin{center}
{\small
\begin{tabular}{lccc}
\hline
 \\[-1.6ex]
\textbf{Method} & \textbf{MMLU} & \textbf{Direct} & \textbf{Ours}\\
 \\[-2ex]
\hline
 \\[-2ex]
 \rowcolor{Gray}
 Prediction LLM &&& \\
Mistral-v0.1-7B & 60.1 & 0.20 & 0.30 \\
OLMo-2-7B & 61.3 & 0.05 & 0.32 \\
Llama-3-8B & 66.6 & \textbf{0.33} & 0.37 \\
Llama-3.1-8B & \textbf{71.3} & 0.32 & \textbf{0.45} \\
 \\[-2ex]
\hline
 \\[-2ex]
\rowcolor{Gray}
Examples LLM &&& \\
Mistral-v0.1-7B & 60.1 & 0.20 & 0.22 \\
OLMo-2-7B & 61.3 & 0.05 & {\color{BrickRed} \textbf{0.20}} \\
Llama-3-8B & 66.6 & \textbf{0.33} & 0.37 \\
Llama-3.1-8B & \textbf{71.3} & 0.32 & \textbf{0.42} \\
 \\[-2ex]
\hline
\end{tabular}
\caption{\textbf{Ablation on LLMs Used for Example Generation and Prediction.} Methods are sorted by ascending MMLU ($5$-shot) \cite{mmlu2020}. All models are their "Instruct" variant. Prompts used for each result are detailed in \cref{sec:appendix:prompt_for_summaries_and_prediction}.}
\label{table:architecture_performance}
}
\end{center}
\end{table}

%% file: appendix.tex
\section{BERTScore and NewsRoom}\label{sec:appendix:bertscore_and_newsroom}

\input{tables/newsroom_issue}
The NewsRoom dataset is often cited in works involving summarization metrics. However, after computing the BERTScore for the subset of NewsRoom's test data that includes human evaluations, we found that BERTScore with \texttt{roberta-large} attained results comparable to those of a direct scoring approach with \texttt{GPT-4-turbo} \cite{newsroom_dataset, liu2024aligning}. Conversely, BERTScore with \texttt{roberta-large} performed markedly worse than \texttt{GPT-4-turbo} on the SummEval annotated dataset, presenting an inconsistency in the expected discrepancies between the results of the methods tested on NewsRoom \cite{liu2024aligning}. 

This is supported by NewsRoom's overall higher correlation between axes, displayed in \cref{tab:newsroom_axis_spearman}, which could result in unusually high scores from BERTScore due to the lack of discernment between the axes in the provided ground-truth human evaluation scores. For comparison, we also include SummEval axis correlations in \cref{tab:summeval_axis_spearman}, whose correlations with the exception of the coherence/relevance pair are lower.

\begin{table}[h]
\centering
\small
\setlength{\tabcolsep}{4pt}
\begin{tabular}{lcccc}
\toprule
           & COH & FLU & CON & REL \\
\midrule
COH    & 1.000 & 0.197 & 0.400 & 0.787 \\
FLU      & 0.197 & 1.000 & 0.317 & 0.254 \\
CON  & 0.400 & 0.317 & 1.000 & 0.458 \\
REL    & 0.787 & 0.254 & 0.458 & 1.000 \\
\bottomrule
\end{tabular}
\caption{\textbf{SummEval Spearman Axis Sample-Level Correlations}. COH, FLU, CON, and REL are the abbreviations of "coherence," "fluency," "consistency," and "relevance," respectively.}
\label{tab:summeval_axis_spearman}
\end{table}

\begin{table}[h]
\centering
\small
\setlength{\tabcolsep}{4pt}
\begin{tabular}{lcccc}
\toprule
                  & COH & FLU & INF & REL \\
\midrule
COH         & 1.000 & 0.788 & 0.753 & 0.559 \\
FLU           & 0.788 & 1.000 & 0.674 & 0.547 \\
INF   & 0.753 & 0.674 & 1.000 & 0.624 \\
REL         & 0.559 & 0.547 & 0.624 & 1.000 \\
\bottomrule
\end{tabular}
\caption{\textbf{NewsRoom Spearman Axis Sample-Level Correlations}. COH, FLU, INF, and REL are the abbreviations of "coherence," "fluency," "informativeness," and "relevance," respectively.}
\label{tab:newsroom_axis_spearman}
\end{table}



\section{System, Sample, and Summary-Level Performance Definitions}\label{sec:appendix:all_performances}
There are three main correlation metrics found within NLG meta-evaluation works: \textit{system-level}, \textit{sample-level}, and \textit{summary-level}. In terms of the SummEval \cite{summeval_dataset} dataset, these correspond to (1) the correlation of the average machine score with the automatic rater versus the human score, (2) the average machine correlation within each document, and (3) the average correlation over all documents after throwing away the machine ID. Formally, for a given correlation metric (e.g. Spearman's $\rho$), a group of abstractive summarization machines of size $M$, and a number of full-texts $T$, we can choose the following metrics: \\

\noindent
\textit{System-Level:}
\begin{align*}
    F^{system} &= \rho(\frac{1}{n}\sum_{i=1}^T\begin{bmatrix}
        pred_{i,1}, \hspace{.2cm}human_{i,1} \\
        ... \\
        pred_{i,M}, human_{i,M}
    \end{bmatrix}) 
\end{align*}
 \\
\noindent
\textit{Sample-Level:}
\begin{align*}
    F^{sample} &= \frac{1}{n}\sum^{T}_{i=1}\rho(\begin{bmatrix}
        pred_{i,1}, human_{i,1} \\
        ... \\
        pred_{i,M}, human_{i,M}
    \end{bmatrix}) 
\end{align*}
 \\
\noindent
\textit{Summary-Level:}
\begin{align*}
    F^{summary} &= \rho(\begin{bmatrix}
        pred_{1,1}, human_{1,1} \\
        ... \\
        pred_{1,M}, human_{1,M} \\ 
        pred_{2,1}, human_{2,1} \\
        ... \\
        pred_{2,M}, human_{2,M} \\ 
        ... \\ 
        pred_{T,1}, human_{T,1} \\
        ... \\
        pred_{T,M}, human_{T,M} \\ 
    \end{bmatrix}) 
\end{align*}

where $pred_{i,j}$ is the prediction for the $j^{th}$ machine's response on the $i^{th}$ document and $human_{i,j}$ is the ground truth label for that machine response.

\section{Prompts Used for Summary Generation and Prediction}\label{sec:appendix:prompt_for_summaries_and_prediction}

Next, we provide the templates for generating synthetic summaries and prediction for different schemes. We organize the section in the following way: 
\begin{enumerate}
    \item SummEval 
    \begin{enumerate}
        \item Best/Worst (Scores 1,5): \cref{tab:prompt_summeval_best_worst}
        \item Recursive (Scores 2,3,4): \cref{tab:prompt_summeval_recursive}
        \item $p(\texttt{"Better"|i}), ...$: \cref{tab:prompt_summeval_predict_bws}
        \item $p(\texttt{"Yes"|i}), p(\texttt{"No"|i})$: \cref{tab:prompt_summeval_predict_yesno}
    \end{enumerate}
    \item TopicalChat 
    \begin{enumerate}
        \item Best/Worst (Scores 1,5): \cref{tab:prompt_topicalchat_best_worst}
        \item Recursive (Scores 2,3,4): \cref{tab:prompt_topicalchat_recursive}
        \item $p(\texttt{"Better"|i}), ...$: \cref{tab:prompt_topicalchat_predict_bws}
        \item $p(\texttt{"Yes"|i}), p(\texttt{"No"|i})$: \cref{tab:prompt_topicalchat_predict_yesno}
    \end{enumerate}
    \item HANNA 
    \begin{enumerate}
        \item Best/Worst (Scores 1,5): \cref{tab:prompt_hanna_best_worst}
        \item Recursive (Scores 2,3,4): \cref{tab:prompt_hanna_recursive}
        \item $p(\texttt{"Better"|i}), ...$: \cref{tab:prompt_hanna_predict_bws}
        \item $p(\texttt{"Yes"|i}), p(\texttt{"No"|i})$: \cref{tab:prompt_hanna_predict_yesno}
    \end{enumerate}
\end{enumerate}

\input{tables/example_prompts_best_worst}
\input{tables/example_prompts_recursive}
\input{tables/prediction_prompts}
\input{tables/prediction_prompts_yesno}

%% file: tables/newsroom_issue.tex
\begin{table}[ht!]
\setlength{\belowcaptionskip}{-10pt}
\setlength{\tabcolsep}{4pt}
\begin{center}
{\small
\begin{tabular}{lccccc}
\hline
 \\[-1.6ex]
\textbf{Method} \hspace{1.3cm} & \textbf{COH} & \textbf{REL} & \textbf{INF} & \textbf{FLU} & \textbf{AVG} \\
 \\[-2ex]
\hline
 \\[-2ex]
\rowcolor{Gray}
Other Metrics &&&&&\\
BERTScore$^1$ (F1) & 0.15 & 0.16 & 0.13 & 0.17 & 0.15 \\
\rowcolor{Red}
BERTScore$^2$ (F1) & \textbf{0.65} & \textbf{0.62} & \textbf{0.69} & \textbf{0.58} & \textbf{0.63} \\
GPTScore & 0.31 & 0.35 & 0.26 & 0.31 & 0.31 \\
 \\[-2ex]
\hline
 \\[-2ex]
\rowcolor{Gray}
Mistral 7B &&&&& \\
\rowcolor{PairwiseScoring}
ZEPO & 0.47 & 0.38 & 0.44 & 0.48 & 0.44 \\
\rowcolor{PairwiseScoring}
PairS-beam & \textbf{0.55} & \textbf{0.53} & \textbf{0.48} & \textbf{0.48} & \textbf{0.51} \\
\rowcolor{DirectScoring}
Direct Scoring & 0.32 & 0.39 & 0.20 & 0.26 & 0.29 \\
\rowcolor{DirectScoring}
G-Eval & 0.36 & 0.36 & 0.24 & 0.39 & 0.34 \\
 \\[-2ex]
\hline
 \\[-2ex]
\rowcolor{Gray}
Llama-3 8B &&&&& \\
\rowcolor{PairwiseScoring}
ZEPO & 0.57 & 0.54 & 0.55 & 0.56 & 0.56 \\
\rowcolor{PairwiseScoring}
PairS-beam & \textbf{0.66} & \textbf{0.66} & \textbf{0.73} & \textbf{0.62} & \textbf{0.67} \\
\rowcolor{DirectScoring}
Direct Scoring & 0.42 & 0.41 & 0.30 & 0.29 & 0.36 \\
\rowcolor{DirectScoring}
G-Eval & 0.38 & 0.34 & 0.26 & 0.26 & 0.31 \\
 \\[-2ex]
\hline
 \\[-2ex]
\rowcolor{Gray}
GPT-4-Turbo &&&&& \\
\rowcolor{PairwiseScoring}
ZEPO & - & - & - & - & - \\
\rowcolor{PairwiseScoring}
PairS-beam & \textbf{0.64} & \textbf{0.61} & \textbf{0.67} & \textbf{0.60} & \textbf{0.63} \\
\rowcolor{DirectScoring}
Direct Scoring & 0.55 & 0.54 & 0.57 & 0.60 & 0.57 \\
\rowcolor{DirectScoring}
G-Eval & 0.58 & 0.55 & 0.57 & 0.58 & 0.57\\
 \\[-2ex]
\hline
\end{tabular}
\caption{\textbf{NewsRoom Performance.}  BERTScore$^1$ refers to using \texttt{bert-base-uncased} whereas $^2$ refers to using \texttt{roberta-large}. The performance of \texttt{roberta-large} is \colorbox{Red}{highlighted in red.} Other numbers are reported from PairS \cite{pairs2024} and ZEPO \cite{zepo}}
\label{table:newsroom_issue}
}
\end{center}
\end{table}

%% file: tables/example_prompts_best_worst.tex
\begin{table*}[h!]
\setlength{\belowcaptionskip}{-10pt}
\setlength{\tabcolsep}{3pt}
\begin{center}
{
\begin{tabular}{p{\linewidth}}
\hline
 \\[-2ex]
\texttt{You will be given a source document and an evaluation dimension for a summary. Your task is to write the \{\{ worst\_best : str \}\ possible summary you can think of with regards to this dimension.}
\newline

\texttt{Your response should only include the \{\{ worst\_best : str \}\} possible summary you can create without any additional text. The summary must be non-empty and directly summarize the article without using phrases like "This article is about."}
\newline

\texttt{Evaluation Criteria:}
\newline

\texttt{\{\{ col\_title : str \}\} - \{\{ col\_description : str \}\}}
\newline

\texttt{Document:}
\newline

\texttt{\{\{ article : str \}\}}
\newline
 \\[-2ex]
\hline
\end{tabular}
\caption{\textbf{SummEval Best/Worst Synthetic Example Prompt Template.}}
\label{tab:prompt_summeval_best_worst}
}
\end{center}
\setlength{\belowcaptionskip}{-10pt}
\setlength{\tabcolsep}{3pt}
\begin{center}
{
\begin{tabular}{p{\linewidth}}
\hline
 \\[-2ex]
\texttt{You will be given a conversation between two people and an evaluation dimension for a response to the most recent message. Your task is to write the \{\{ worst\_best : str \}\} possible response you can think of with regards to this dimension.}
\newline

\texttt{Your response should exactly be the \{\{ worst\_best : str \}\} possible response without any additional text. The response must be non-empty. Try to make the response less than a few sentences and keep the tone informal.}
\newline

\texttt{Evaluation Criteria:}

\texttt{\{\{ col\_title : str \}\} - \{\{ col\_description : str \}\}}
\newline

\texttt{Conversation:}

\texttt{\{\{ context : str \}\}}
\newline
 \\[-2ex]
\hline
\end{tabular}
\caption{\textbf{TopicalChat Best/Worst Synthetic Example Prompt Template.}}
\label{tab:prompt_topicalchat_best_worst}
}
\end{center}
\setlength{\belowcaptionskip}{-10pt}
\setlength{\tabcolsep}{3pt}
\begin{center}
{
\begin{tabular}{p{\linewidth}}
\hline
 \\[-2ex]
\texttt{You will be given an idea for a story and an evaluation dimension for that story. Your task is to write the \{\{ worst\_best : str \}\} possible story you can think of with regards to this dimension.}
\newline

\texttt{Your response should exactly be the \{\{ worst\_best : str \}\} possible story without any additional text. The summary must be non-empty and directly summarize the article without using phrases like "This story is about." Try to keep the story less than 150 words and end on a full sentence. Don't use paragraphs.}
\newline

\texttt{Evaluation Criteria:}

\texttt{\{\{ col\_title : str \}\} - \{\{ col\_description : str \}\}}
\newline

\texttt{Story Idea:}

\texttt{\{\{ story\_prompt : str \}\}}
\newline
 \\[-2ex]
\hline
\end{tabular}
\caption{\textbf{HANNA Best/Worst Synthetic Example Prompt Template.}}
\label{tab:prompt_hanna_best_worst}
}
\end{center}
\end{table*}

%% file: tables/example_prompts_recursive.tex
\begin{table*}[h!]
\setlength{\belowcaptionskip}{-10pt}
\setlength{\tabcolsep}{3pt}
\begin{center}
{\small
\begin{tabular}{p{\linewidth}}
\hline
 \\[-2ex]
\texttt{You will be given a source document and an evaluation dimension for a summary. Your task is to write a summary which is higher quality than one summary (Bad Summary) but lower quality than another (Good Summary).}
\newline

\texttt{Your response should only include the in-between summary without any additional text. The summary must be non-empty and directly summarize the article without using phrases like "This article is about."}
\newline

\texttt{Evaluation Criteria:}

\texttt{\{\{ col\_title : str \}\} - \{\{ col\_description : str \}\}}
\newline

\texttt{Bad Summary:}

\texttt{\{\{ worse\_summary : str \}\}}
\newline

\texttt{Good Summary:}

\texttt{\{\{ better\_summary : str \}\}}
\newline

\texttt{Document:}

\texttt{\{\{ article : str \}\}}
\newline
 \\[-2ex]
\hline
\end{tabular}
\caption{\textbf{SummEval Recursive Synthetic Example Prompt Template.}}
\label{tab:prompt_summeval_recursive}
}
\end{center}
\setlength{\belowcaptionskip}{-10pt}
\setlength{\tabcolsep}{3pt}
\begin{center}
{\small
\begin{tabular}{p{\linewidth}}
\hline
 \\[-2ex]
\texttt{You will be given a conversation between two people and an evaluation dimension for a response to the most recent message. Your task is to write a response which is higher quality than one response (Bad Response) but lower quality than another (Good Response).}
\newline

\texttt{Your message should exactly be the in-between response without any additional text. The response must be non-empty. Try to make the response less than a few sentences and keep the tone informal.}
\newline

\texttt{Evaluation Criteria:}

\texttt{\{\{ col\_title : str \}\} - \{\{ col\_description : str \}\}}
\newline

\texttt{Conversation:}

\texttt{\{\{ context : str \}\}}
\newline

\texttt{Bad Response:}

\texttt{\{\{ worse\_summary : str \}\}}
\newline

\texttt{Good Response:}

\texttt{\{\{ better\_summary : str \}\}}
\newline
 \\[-2ex]
\hline
\end{tabular}
\caption{\textbf{TopicalChat Recursive Synthetic Example Prompt Template.}}
\label{tab:prompt_topicalchat_recursive}
}
\end{center}
\setlength{\belowcaptionskip}{-10pt}
\setlength{\tabcolsep}{3pt}
\begin{center}
{\small
\begin{tabular}{p{\linewidth}}
\hline
 \\[-2ex]
\texttt{You will be given an idea for a story and an evaluation dimension for a story. Your task is to write a story which is higher quality than one story (Bad Story) but lower quality than another (Good Story).}
\newline

\texttt{Your response should exactly be the in-between story without any additional text. The story must be non-empty, be related to the idea, and not use phrases like "This story is about." Try to keep the story less than 150 words and end on a full sentence. Do not use paragraphs.}
\newline

\texttt{Evaluation Criteria:}

\texttt{\{\{ col\_title : str \}\} - \{\{ col\_description : str \}\}}
\newline

\texttt{Story Idea:}

\texttt{\{\{ story\_prompt : str \}\}}
\newline

\texttt{Bad Story:}

\texttt{\{\{ worse\_summary : str \}\}}
\newline

\texttt{Good Story:}

\texttt{\{\{ better\_summary : str \}\}}
\newline
 \\[-2ex]
\hline
\end{tabular}
\caption{\textbf{HANNA Recursive Synthetic Example Prompt Template.}}
\label{tab:prompt_hanna_recursive}
}
\end{center}
\end{table*}

%% file: tables/prediction_prompts.tex
\begin{table*}[h!]
\setlength{\belowcaptionskip}{-10pt}
\setlength{\tabcolsep}{3pt}
\begin{center}
{\scriptsize
\begin{tabular}{p{\linewidth}}
\hline
 \\[-2ex]
\texttt{You will be given a news article, one target summary of that news article, and a reference summary of that article. Your goal is to say whether the quality of the target summary is better, worse, or similar to the reference summary with respect to \{\{ col : str \}\}.}
\newline

\texttt{Evaluation Criteria:}

\texttt{\{\{ col\_title \}\}: \{\{ col\_description \}\}}
\newline

\texttt{Evaluation Steps:}

\texttt{1. Read the news article carefully and identify the main facts and details it presents.}

\texttt{2. Read the target summary and example summary. Compare them to the article.}

\texttt{3. Compare the quality of the target summary to reference summary with respect to \{\{ col : str \}\}.}

\texttt{4. Respond with only one of the following: "Better" "Worse" or "Similar" which indicate whether the target summary is better than, worse than, or similar to the reference summary.}
\newline

\texttt{Original Article:}

\texttt{\{\{ article : str \}\}}
\newline

\texttt{Reference Summary:}

\texttt{\{\{ icl\_summary : str \}\}}
\newline 

\texttt{Target Summary:}

\texttt{\{\{ target\_summary : str \}\}}
\newline
 \\[-2ex]
\hline
\end{tabular}
\caption{\textbf{SummEval $p(\texttt{"Better"}), ...$ Prediction Prompt Template.}}
\label{tab:prompt_summeval_predict_bws}
}
\end{center}
\setlength{\belowcaptionskip}{-10pt}
\setlength{\tabcolsep}{3pt}
\begin{center}
{\scriptsize
\begin{tabular}{p{\linewidth}}
\hline
 \\[-2ex]
\texttt{You will be given a news article, one target summary of that news article, and a reference summary of that article. Your goal is to say whether the quality of the target summary is better, worse, or similar to the reference summary with respect to \{\{ col : str \}\}.}
\newline

\texttt{Evaluation Criteria:}

\texttt{\{\{ col\_title \}\}: \{\{ col\_description \}\}}
\newline

\texttt{Evaluation Steps:}

\texttt{1. Read the story carefully.}

\texttt{2. Read the reference story and evaluation story. Compare them to the idea.} 

\texttt{3. Compare the quality of the evaluation story to the reference story with respect to \{\{ col : str \}\}.}

\texttt{4. Respond with only one of the following: "Better" "Worse" or "Similar" which indicate whether the evaluation story is better than, worse than, or similar to the reference story.}
\newline

\texttt{Story Idea:}

\texttt{\{\{ story\_prompt : str \}\}}
\newline

\texttt{Reference Story:}

\texttt{\{\{ icl\_summary : str \}\}}
\newline

\texttt{Evaluation Story:}

\texttt{\{\{ target\_summary : str \}\}}
\newline
 \\[-2ex]
\hline
\end{tabular}
\caption{\textbf{TopicalChat $p(\texttt{"Better"}), ...$ Prediction Prompt Template.}}
\label{tab:prompt_topicalchat_predict_bws}
}
\end{center}
\setlength{\belowcaptionskip}{-10pt}
\setlength{\tabcolsep}{3pt}
\begin{center}
{\scriptsize
\begin{tabular}{p{\linewidth}}
\hline
 \\[-2ex]
\texttt{You will be given a news article, one target summary of that news article, and a reference summary of that article. Your goal is to say whether the quality of the target summary is better, worse, or similar to the reference summary with respect to \{\{ col : str \}\}.}
\newline

\texttt{Evaluation Criteria:}

\texttt{\{\{ col\_title \}\}: \{\{ col\_description \}\}}
\newline

\texttt{Evaluation Steps:}

\texttt{1. Read the story carefully.}

\texttt{2. Read the reference story and evaluation story. Compare them to the idea.} 

\texttt{3. Compare the quality of the evaluation story to the reference story with respect to \{\{ col : str \}\}.}

\texttt{4. Respond with only one of the following: "Better" "Worse" or "Similar" which indicate whether the evaluation story is better than, worse than, or similar to the reference story.}
\newline

\texttt{Story Idea:}

\texttt{\{\{ story\_prompt : str \}\}}
\newline

\texttt{Reference Story:}

\texttt{\{\{ icl\_summary : str \}\}}
\newline

\texttt{Evaluation Story:}

\texttt{\{\{ target\_summary : str \}\}}
\newline
 \\[-2ex]
\hline
\end{tabular}
\caption{\textbf{HANNA $p(\texttt{"Better"}), ...$ Prediction Prompt Template.}}
\label{tab:prompt_hanna_predict_bws}
}
\end{center}
\end{table*}

%% file: tables/prediction_prompts_yesno.tex
\begin{table*}[h!]
\setlength{\belowcaptionskip}{-10pt}
\setlength{\tabcolsep}{3pt}
\begin{center}
{
\begin{tabular}{p{\linewidth}}
\hline
 \\[-2ex]
\texttt{Here is a news article:}

\texttt{\{\{ article : str \}\}}
\newline

\texttt{Summary 1:}

\texttt{\{\{ target\_summary : str \}\}}
\newline

\texttt{Summary 2:}

\texttt{\{\{ icl\_summary : str \}\}}
\newline

\texttt{Does Summary 1 \{\{ prediction : str \}\} than Summary 2?}
\newline

\texttt{Respond with only one of the following: "Yes" or "No".}
\newline
 \\[-2ex]
\hline
\end{tabular}
\caption{\textbf{SummEval $p(\texttt{"Yes"}), p(\texttt{"No"})$ Prediction Prompt Template.}}
\label{tab:prompt_summeval_predict_yesno}
}
\end{center}
\setlength{\belowcaptionskip}{-10pt}
\setlength{\tabcolsep}{3pt}
\begin{center}
{
\begin{tabular}{p{\linewidth}}
\hline
 \\[-2ex]
\texttt{Here is a conversation:}

\texttt{\{\{ context : str \}\}}
\newline

\texttt{Response 1:}

\texttt{\{\{ target\_summary : str \}\}}
\newline

\texttt{Response 2:}

\texttt{\{\{ icl\_summary : str \}\}}
\newline

\texttt{Does Response 1 \{\{ prediction : str \}\} than Response 2?}
\newline

\texttt{Respond with only one of the following: "Yes" or "No".}
\newline
 \\[-2ex]
\hline
\end{tabular}
\caption{\textbf{TopicalChat $p(\texttt{"Yes"}), p(\texttt{"No"})$ Prediction Prompt Template.}}
\label{tab:prompt_topicalchat_predict_yesno}
}
\end{center}
\setlength{\belowcaptionskip}{-10pt}
\setlength{\tabcolsep}{3pt}
\begin{center}
{
\begin{tabular}{p{\linewidth}}
\hline
 \\[-2ex]
\texttt{Story 1:}

\texttt{\{\{ target\_summary : str \}\}}
\newline

\texttt{Story 2:}

\texttt{\{\{ icl\_summary : str \}\}}
\newline

\texttt{Does Story 1 \{\{ prediction : str \}\} than Story 2?}
\newline

\texttt{Respond with only one of the following: "Yes" or "No".}
\newline
 \\[-2ex]
\hline
\end{tabular}
\caption{\textbf{HANNA $p(\texttt{"Yes"}), p(\texttt{"No"})$ Prediction Prompt Template.}}
\label{tab:prompt_hanna_predict_yesno}
}
\end{center}
\end{table*}